\documentclass[nohyperref]{article}
\PassOptionsToPackage{sort,numbers,compress}{natbib}
\usepackage{amsmath}
\usepackage[utf8]{inputenc} 
\usepackage[T1]{fontenc}    
\usepackage[normalem]{ulem}
\usepackage{amsthm, amssymb, thmtools, thm-restate}
\usepackage{bbm}
\usepackage[authoryear,sort]{natbib}
\usepackage[margin=1in]{geometry}
\usepackage{minitoc}
\usepackage{algorithm}
\usepackage{algorithmicx}
\usepackage{algpseudocode}
\usepackage{subfigure}
\usepackage{nicefrac}
\usepackage{booktabs}       
\usepackage{amsfonts}       
\usepackage{nicefrac}       
\usepackage{microtype}      
\usepackage{xcolor}         
\usepackage{url}            
\usepackage{hyperref}
\usepackage{enumitem}
\usepackage{authblk}
\usepackage[capitalize,noabbrev]{cleveref}
\usepackage{caption}
\usepackage{tikz}
\usetikzlibrary{fit,calc}

\colorlet{pink}{red!40}
\colorlet{green}{cyan!60}

\def\*#1{\boldsymbol{#1}}

\newtheorem{theorem}{Theorem}

\crefname{assumption}{Assumption}{Assumptions}

\crefname{example}{Example}{Examples}

\crefname{thm}{Theorem}{Theorems}

\crefname{lem}{Lemma}{Lemmas}

\crefname{prop}{Proposition}{Propositions}

\crefname{cor}{Corollary}{Corollaries}

\newcommand{\step}{\textbf{STEP}}
\newcommand{\as}{\textbf{AutoSwitch}}

\hypersetup{
    colorlinks,
    linkcolor={blue!50!black},
    citecolor={blue!50!black},
    urlcolor={blue!80!black}
}

\title{{\step}: Learning N:M Structured Sparsity Masks from Scratch with Precondition}
\allowdisplaybreaks

\author[1]{Yucheng Lu\thanks{Corresponds to: yl2967@cornell.edu.}}
\author[2]{Shivani Agrawal}
\author[2]{Suvinay Subramanian}
\author[2]{Oleg Rybakov}
\author[1]{\\ Christopher De Sa}
\author[2]{Amir Yazdanbakhsh}
\affil[1]{Department of Computer Science, Cornell\ University}
\affil[2]{Google Research}
\date{}

\begin{document}

\maketitle

\begin{abstract}
    Recent innovations on hardware (e.g. Nvidia A100) have motivated learning N:M structured sparsity masks from scratch for fast model inference. However, state-of-the-art learning recipes in this regime (e.g. SR-STE) are proposed for non-adaptive optimizers like momentum SGD, while incurring non-trivial accuracy drop for Adam-trained models like attention-based LLMs.
    In this paper, we first demonstrate such gap origins from poorly estimated second moment (i.e. variance) in Adam states given by the masked weights. We conjecture that learning N:M masks with Adam should take the critical regime of variance estimation into account. In light of this, we propose {\step}, an Adam-aware recipe that learns N:M masks with two phases: first, {\step} calculates a reliable variance estimate (\emph{precondition phase}) and subsequently, the variance remains fixed and is used as a precondition to learn N:M masks (\emph{mask-learning phase}). {\step} automatically identifies the switching point of two phases by dynamically sampling variance changes over the training trajectory and testing the sample concentration. Empirically, we evaluate {\step} and other baselines such as ASP and SR-STE on multiple tasks including CIFAR classification, machine translation and LLM fine-tuning (BERT-Base, GPT-2). We show {\step} mitigates the accuracy drop of baseline recipes and is robust to aggressive structured sparsity ratios.
\end{abstract}
\section{Introduction}
\label{sec:introduction}
Overparameterized Deep Neural Networks (DNNs) have shown promising performance on various applications, such as language modeling \citep{brown2020language}, translation \citep{vaswani2017attention} and image classification \citep{liu2021swin}. However, modern DNNs usually contain millions of billions of parameters (e.g. BERT \citep{devlin2018bert} and GPT \citep{brown2020language}), which hinders the inference scalability. Recent innovation on hardware architecture suggests structured sparsity is a promising way of alleviating this issue by deploying N:M masks during inference (N out of consecutive M elements in the the weight tensor are kept while others are pruned). N:M masks accelerate model inference with regular sparse structures \citep{pool2020accelerating,fang2022algorithm}. 
Compared to traditional unstructured sparsity \citep{frankle2018lottery, lee2018snip, evci2020rigging} or channel/block structured sparsity algorithms \citep{wen2016learning,li2016pruning,he2017channel},
adopting N:M masks has negligible evaluation degradation and progressively co-design algorithm (sparse matrix multiplication) and hardware (e.g. Nvidia Ampere Sparse Tensor Core), reaching a desirable trade-off.

Following this line of research, recent studies indicate it is critical (and also possible) to learn these N:M masks from scratch, without additional training or finetuning steps.
Representative methods in this domain include SR-STE \citep{zhou2021learning}, DominoSearch \citep{sun2021dominosearch} and Decaying Mask \citep{kao2022training}, which sparsify the model weights during each forward pass in training to compute gradients, and update them to models. While these methods demonstrate promising results with momentum SGD, their performance over adaptive optimizers, such as Adam, is less satisfactory (Section~\ref{sec:preliminary}). 
This implies the benefits of sparsity are largely traded-off by adaptivity in training, leading to slow convergence on many state-of-the-art models \citep{zhang2020adaptive}.
In light of this, in this paper we answer the question:
\begin{center}\emph{Can we learn N:M structured sparsity masks with Adam, without model degradation?}\end{center}
Motivated by the insights from recent studies on critical learning regime of Adam in a distributed learning environment \cite{tang20211,lu2022maximizing}, we first hypothesize that with masked weights, the back propagation leads to noisy gradients and gives a poorly estimated variance (running average of second moment gradients) in the Adam states. It essentially breaks the proper scaling of the coordinate-wise learning rate.

To alleviate this, we propose {\step} that learns N:M masks with two phases: 1) in the first phase, no mask is applied and {\step} explores the gradient space to obtain a reliable variance estimate (\emph{precondition phase}); 2) in the second phase, such estimate remains fixed and is used to learn N:M masks (\emph{mask-learning phase}).
While previous works have had similar ideas on two-phase training paradigm under the context of low-precision training \citep{tang2020apmsqueeze,tang20211,lu2022maximizing}, the switching point of two phases is still decided by heuristics or redundant hyperparameter tuning. In contrast, {\step} leverages a novel {\as} subroutine that samples the variance update along the training trajectory and tests their concentration.

Our contributions in this paper can be summarized as follows:
\begin{itemize}[nosep,leftmargin=12pt]
    \item We introduce {\step}, a recipe for learning N:M structured sparsity masks from scratch with Adam. {\step} addresses the accuracy drop of state-of-the-art recipes (e.g. SR-STE) with Adam. {\step} involves a novel subroutine named {\as}, which automatically separates the training into precondition and mask learning phases by dynamically testing variance concentration.
\item We provide in-depth analysis on why using preconditioning in Adam is justifiable, and prove in theory that under the same conditions given in original Adam paper \citep{kingma2014adam}, the precondition error from {\step} remains bounded and the averaged accumulated approximation error is decreasing over time. 
    \item We perform extensive experiments on CIFAR image classification, WMT machine translation, fine-tuning BERT on GLUE and GPT-2 on WikiText-2/-103 that {\step} mitigates the accuracy drop of baseline algorithms, and is robust to aggressive structured sparsity ratios.
\end{itemize}

\section{Related Work}
\label{sec:relatedwork}

\paragraph{Recipes for Learning N:M Structured Sparsity Masks from Scratch.}
With the proposition of Sparse Tensor Cores introduced in the NVIDIA Ampere GPU
architecture \citep{mishra2021accelerating}, there has been an increasing interest of learning N:M structured sparsity masks from scratch. \citet{zhou2021learning} initiatively proposes SR-STE that leverages sparse refinement when evaluating gradients via masked weights (termed Straight Through Estimator). Subsequently, \citet{sun2021dominosearch} and \citet{kao2022training} extend SR-STE towards using adaptive N:M ratios across layers and steps.
While these works focus on learning the N:M masks from scratch, other works have separate discussions. For instance,
\citet{holmes2021nxmtransformer} proposes a general framework to learn the structured sparsity mask on a  pre-trained model specifically.
\citet{hubara2021accelerated} aims to find N:M masks to speed up training rather than inference.
\citet{pool2021channel} advocates a pre-permutation yields better results for N:M sparsity and \citet{chmiel2022optimal} discusses the structured sparsity on activations.

\paragraph{Critical Learning Regime for Adam Variance.}
The existence of a critical learning regime during neural network training has been observed by various studies \citep{achille2018critical,frankle2018lottery,gur2018gradient}. Many prior works including \citep{jastrzkebski2018relation,jastrzebski2020break} highlight that the early phase of training with SGD determines the difficulty of entire training.
Lately, studies including \citep{agarwal2021adaptive,tang2020apmsqueeze,tang20211} suggests the critical learning regime also exists for Adam-type optimizers \citep{kingma2014adam} in a distributed learning environment.
More specifically, it has been pointed out that if we wish to use communication quantization for distributed Adam, then we must run dense Adam for the first few iterations to obtain a reliable variance, followed by iterations where quantization is actually applied \citep{tang2020apmsqueeze,tang20211,li20211,lu2022maximizing}.
Despite the similarity in heuristics to our works, accurately identifying the critical learning regime (i.e. precondition phase) is much more crucial in learning N:M masks: early exiting the precondition phase could lead to unreliable variance estimate while late exit could result in poorly-trained N:M masks. This makes the previous methods on hand-picking the phase length for preconditioning highly unreliable.

\section{Preliminary}

\begin{figure*}[t!]
  \centering
  \subfigure[ResNet18 on CIFAR10]{
    \includegraphics[width=0.46\textwidth]{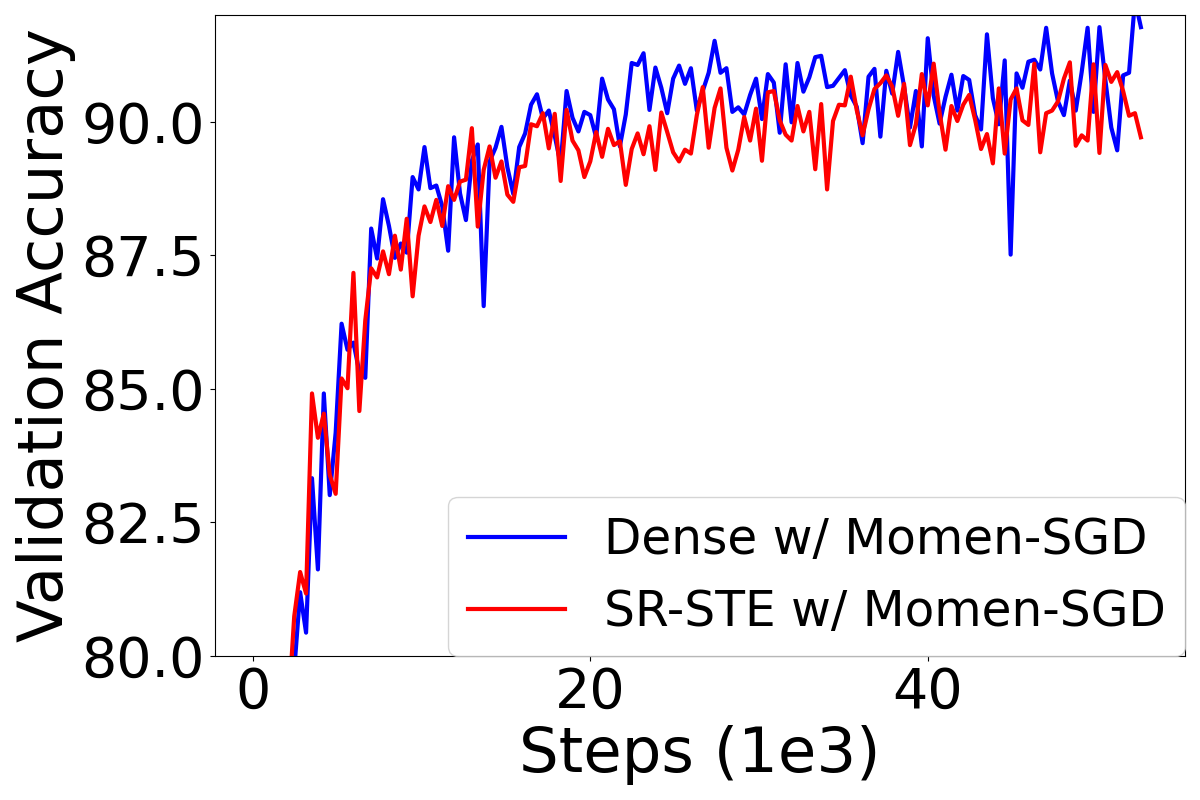}
    \includegraphics[width=0.46\textwidth]{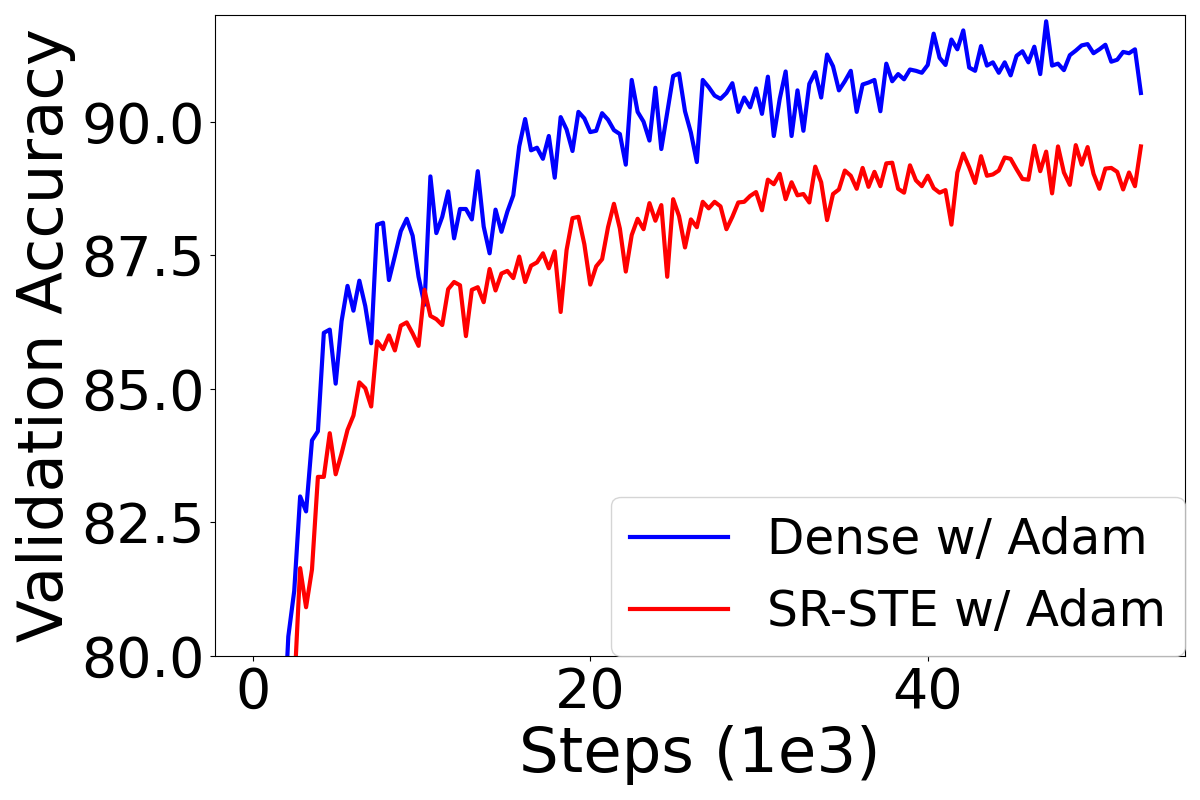}
  }
  \subfigure[DenseNet121 on CIFAR100]{
    \includegraphics[width=0.46\textwidth]{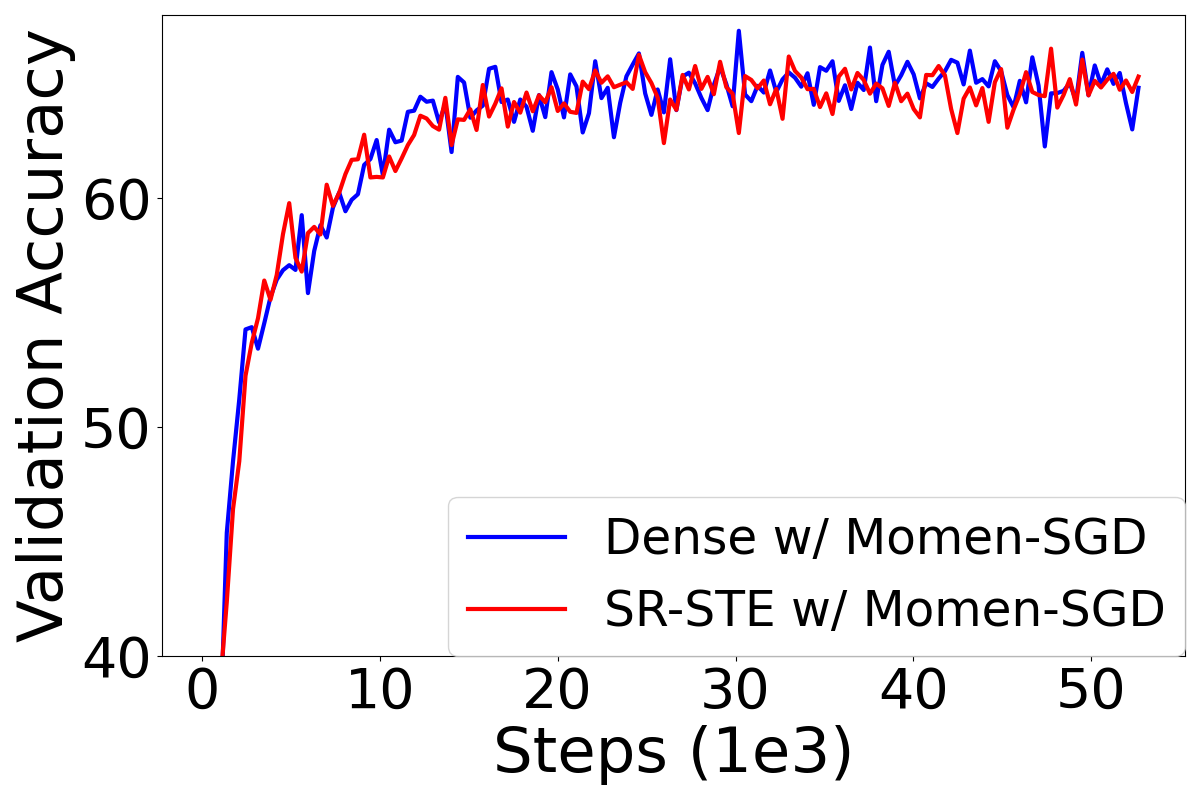}
    \includegraphics[width=0.46\textwidth]{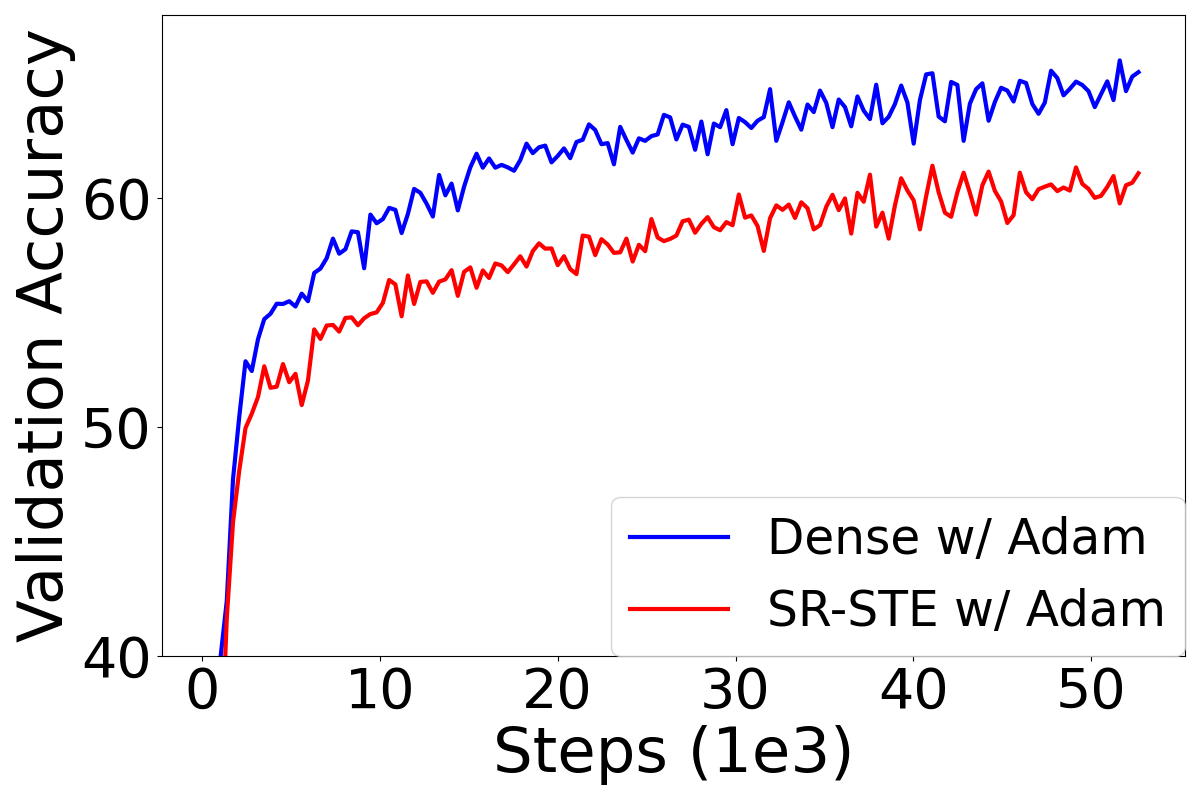}
  }
  \caption{Figures demonstrating the state-of-the-art N:M masks learning recipe SR-STE \citep{zhou2021learning} works with momentum SGD but fails to reach target accuracy when trained with Adam on CIFAR classification tasks. In this demonstration, 1:4 (N=1, M=4) sparsity is applied on all the model weights using the exact implementation from \citep{zhou2021learning}. Note that here we are not comparing the performance between momentum SGD and Adam, but rather focus on the accuracy gap between dense and SR-STE under two different optimizers.}
  \label{fig:srste_issue_demo}
\end{figure*}

\label{sec:preliminary}
In this section, we give a more formal description on the problem formulation. We first provide an overview on the Adam updates and fundamentals to learn N:M masks from scratch with Straight Through Estimator (STE). We also introduce our main baseline SR-STE \citep{zhou2021learning}, the state-of-the-art recipe to learn N:M masks. We conclude this section by showing naively applying SR-STE over Adam incurs non-trivial accuracy drop when training ResNet18 on CIFAR10 \citep{he2016deep} and DenseNet121 on CIFAR100 \citep{huang2017densely}.

\paragraph{Overview of Adam Updates.}
Model training in general can be formulated as an optimization problem, i.e., finding a set of target model weights $\*w^*\in\mathbb{R}^d$ that minimizes the loss function:
\begin{align}
    \*w^* = \arg\min_{\*w\in\mathbb{R}^d} \left[f(\*w) = \mathbb{E}_{\zeta\sim\mathcal{D}}f(\*w;\zeta) \right].
\end{align}
where $\mathcal{D}$ denotes the training set and $f(\*w;\zeta)$ is the loss incurred over sample $\zeta$ given $d$-dimensional model parameters $\*w$. The Adam optimizer \citep{kingma2014adam} solves this problem iteratively with an adaptive learning rate schedule. Concretely, with some initialized value $\*w_1$, for any $t\geq 1$, the update formula of Adam\footnote{Note that in Adam, operations like division should act element-wise.} can be summarized as:
\begin{align}
\label{equ:adam_grad}
\text{(Sample Gradient) } & \*g_t = \nabla f(\*w_t;\zeta_t), \hspace{1em} \zeta_t\sim\mathcal{D}, \\
    \text{(Update $\*m$) } & \*m_{t+1} =  \beta_1\*m_{t} + (1-\beta_1)\*g_t, \\
\text{(Update $\*v$) } & \*v_{t+1} = \beta_2\*v_{t} + (1-\beta_2)(\*g_t)^2, \\
    \text{(Correct Bias) } & \hat{\*m}_{t+1} = \frac{\*m_{t+1}}{1-\beta_1^t}, \\
\text{(Correct Bias) } & \hat{\*v}_{t+1} = \frac{\*v_{t+1}}{1-\beta_2^t}, \\
    \text{(Update Model) } & \*w_{t+1} = \*w_t - \underbrace{\frac{\gamma_t}{\sqrt{\hat{\*v}_{t+1} + \epsilon}}}_{\text{adaptive learning rate}} \odot \hat{\*m}_{t+1},
\end{align}
where $\gamma_t$ is the learning rate at step $t$, $\epsilon$ is a small constant to prevent zero division, $\beta_1$ and $\beta_2$ are tunable decaying factors. 
The running average of first and second gradient moments $\*m$ and $\*v$ are usually referred to as \emph{momentum} and \emph{variance}, respectively.
The Adam optimizer (and its variants) has been adopted as the folklore method to train many models since its proposition. In recent studies like \citep{zhang2020adaptive}, it has been found that Adam is critical for many attention-based foundation models to achieve state-of-the-art model quality.

\paragraph{Overview of SR-STE.}
Learning N:M structured sparsity masks from scratch refers to generating a set of N:M masks at the end of model training, without any additional training steps, and apply these masks during inference.
STE \citep{bengio2013estimating} is a basic method to solve this problem by directly masking the model weights during forward passes, making the gradients mask-aware. This can be formally expressed as: $\forall t \geq 1$
\begin{align}
\label{equ:ste_grad}
    \*g_t = \nabla f(\Pi_t \odot \*w_t; \zeta_t), 
\end{align}
where $\Pi_t$ is an N:M mask obtained based on the magnitude of $\*w_t$.
Comparing Equation~(\ref{equ:adam_grad}) and Equation~(\ref{equ:ste_grad}), the main difference in STE is that the gradient is now computed on the masked weights, while the mask is $\*w_t$ specific at any training step $t$.

Based on STE, SR-STE \citep{zhou2021learning} advocates a regularized version of gradients with masking. Specifically, with a given regularizing coefficient $\lambda$, SR-STE estimates the gradient as:
\begin{align}
    \*g_t = \nabla f(\Pi_t \odot \*w_t; \zeta_t) + \lambda(\*1-\Pi_t) \odot \*w_t,
\end{align}
where $\*1$ denotes all-one vector in $\mathbb{R}^d$. It has been shown in \citep{zhou2021learning} that proper refinement and a well-tuned $\lambda$ mitigates the accuracy drop of momentum SGD over plain STE.

\paragraph{Issue on SR-STE with Adam.}

While the majority of results shown in \citep{zhou2021learning} demonstrates the effectiveness of SR-STE over momentum SGD, here we identify even on simple CIFAR tasks, SR-STE could lead to unsatisfactory sparse models when trained with Adam. We plot the results in Figure~\ref{fig:srste_issue_demo}, which compares the performance of dense training and SR-STE on two models (ResNet18 and DenseNet121) on CIFAR10/100 datasets. We observe that when training a model with Adam, the masks learned by SR-STE incur non-trivial accuracy drop during model inference.

\begin{algorithm}[t!]
	\caption{Proposed {\step} Algorithm}\label{algo:STEP}
	\begin{algorithmic}[1]
		\Require Initial time step $t=0$, initialized model weights $\*w_0$, Adam-related hyperparameters: \{$(\beta_1, \beta_2)$, $\epsilon$ for preventing zero division, initialized momentum and variance $\*m_0=\*0$, $\*v_0=\*0$\}.
		\While{True}
		    \State Sample the data batch $\zeta_t$.
		    \State Compute stochastic gradient $\*g_t=\nabla f(\*w_t;\zeta_t)$.
		    \State Update the momentum: $\*m_{t+1} = \beta_1\*m_{t} + (1-\beta_1)\*g_{t}$.
		    \State Update the variance: $\*v_{t+1} = \beta_2\*v_{t} + (1-\beta_2)(\*g_{t})^2$.
		    \State Correct momentum bias: $\hat{\*m}_{t+1} = \*m_{t+1} / (1-\beta_1^t)$.
		    \State Correct variance bias: $\hat{\*v}_{t+1} = \*v_{t+1} / (1-\beta_2^t)$.
		    \State Update the weights: $\*w_{t+1} = \*w_t - \gamma_t\hat{\*m}_{t+1} / \sqrt{\hat{\*v}_{t+1} + \epsilon}$.
		    \State Update the time $t = t + 1$.
		    \If{$t$ is the switching point}
		        \State Set the preconditioned variance $\*v^* = \*v_t$ and \textbf{break}.
		    \EndIf
		\EndWhile
		\While{$t < T$}
		    \State Sample the data batch $\zeta_t$.
		    \State Compute N:M mask $\Pi_t$ based on the current weights $\*w_t$.
		    \State Compute stochastic gradient $\*g_t=\nabla f(\Pi_t \odot \*w_t;\zeta_t)$.
		    \State Update the momentum: $\*m_{t+1} = \beta_1\*m_{t} + (1-\beta_1)\*g_{t}$.
		    \State Correct momentum bias: $\hat{\*m}_{t+1} = \*m_{t+1} / (1-\beta_1^t)$.
		    \State Update the weights: $\*w_{t+1} = \*w_t - \gamma_t\hat{\*m}_{t+1} / \sqrt{\*v^* + \epsilon}$.
		    \State Update the time $t = t + 1$.
		\EndWhile
		\State Compute N:M mask $\Pi_T$ based on the current weights $\*w_T$.
		\State \textbf{return} $\Pi_T \odot \*w_T$ for inference.
	\end{algorithmic}
\end{algorithm}

\section{{\step}: STE with Precondition}

In this section, we introduce the approach of addressing the aforementioned issue of SR-STE with Adam. The intuition of our method is based on the observation on variance change during model training. We then justify our approach with theory under the same condition in \citep{kingma2014adam}, and illustrate its practicality.

\paragraph{A Closer Look at Variance Change.}

\begin{figure}[t!]
  \centering
  \subfigure[ResNet18 on CIFAR10]{
    \includegraphics[width=0.46\textwidth]{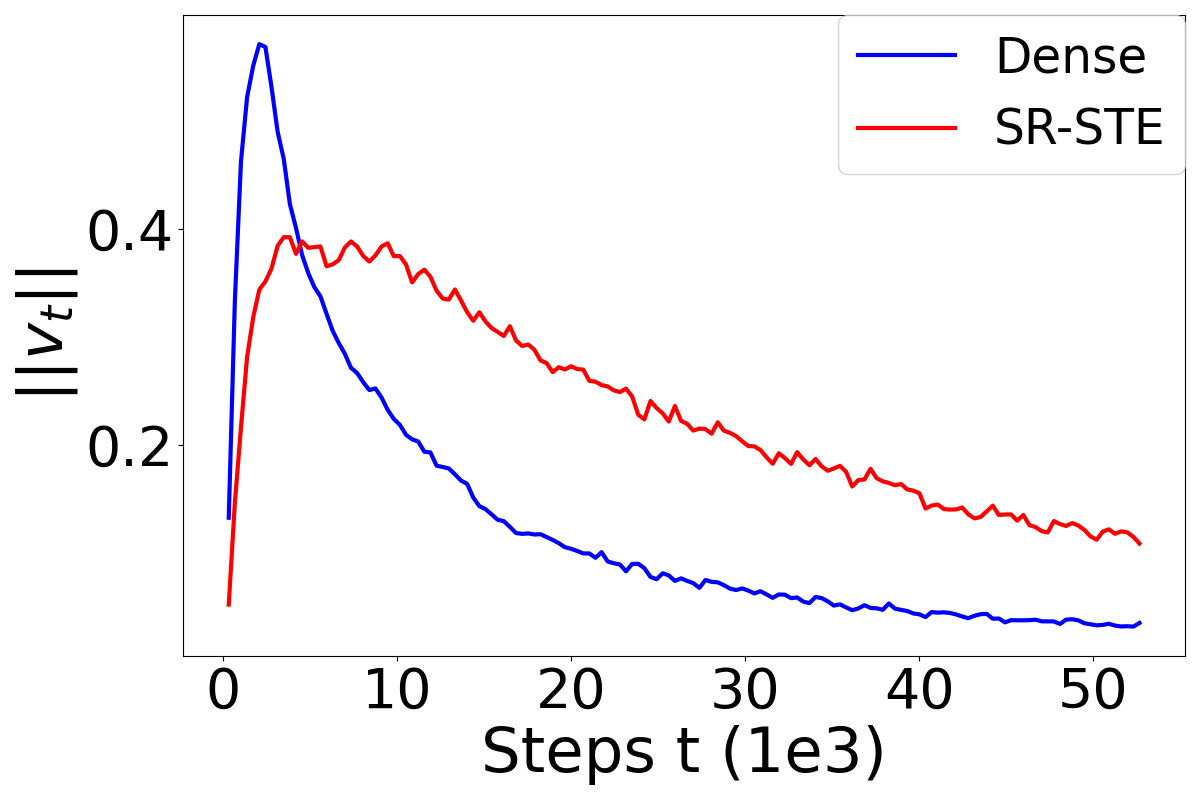}
  }
  \subfigure[DenseNet121 on CIFAR100]{
    \includegraphics[width=0.46\textwidth]{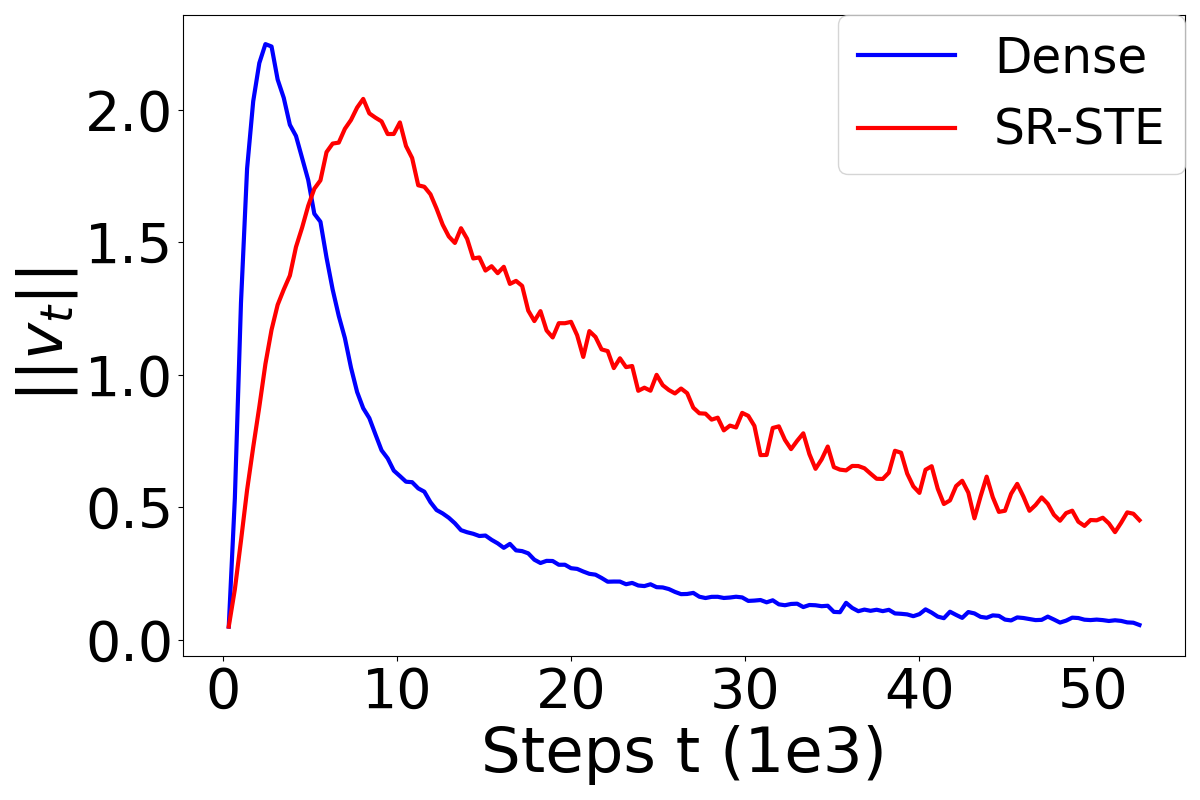}
  }
  \caption{Figure showing variance $\*v_t$ (running average of second moment) change in the Adam states, in the CIFAR tasks shown in Figure~\ref{fig:srste_issue_demo}. In dense training, the variance gradually becomes small in magnitude, which suggests the model converges. In contrast, in SR-STE, the variance norm remains large, which suggests the gradients are noisy even in later stage of the training, and thus it scales down the adaptive learning rates.}
  \label{fig:srste_issue_demo_var_change}
\end{figure}

Motivated by the recent studies on distributed Adam \citep{li20211,tang20211,lu2022maximizing}, we take a closer look at the variance change in the previous tasks and plot them in Figure~\ref{fig:srste_issue_demo_var_change}. We observe that while in both dense training and SR-STE, the variance norm first increases and then decreases, the norm in SR-STE remains large at later stage of learning. This implies the noise obtained in the gradients remains large and essentially scales down the learning rate \citep{kingma2014adam}.

This motivates us to think extensively on the previous success in distributed learning: can we first run dense Adam to obtain a reliable variance, and then learn the N:M masks over the preconditioned variance? While this is mainly based on heuristics in previous works, we next illustrate it is well-justified in theory.

\paragraph{Theoretical Motivation.}
To motivate preconditioned variance, we start from the original objective of having a variance scaler on the learning rate.
In the original Adam paper \citep{kingma2014adam}, it is shown that $\*v_t$ is advocated to capture the expectation of the gradient magnitude at step $t$. In fact, \citet{kingma2014adam} provably shows that if the gradient square $\*g_t^2$ is stationary, i.e. $\mathbb{E}[\*g_i^2]=\mathbb{E}[\*g_j^2]$ for any $i$ and $j$, then $\mathbb{E}[\hat{\*v}_t] = \mathbb{E}[\*g_t^2]$ so that $\hat{\*v}_t$ can be used as an estimator for $\*g_t^2$. Following this intuition, we next prove that under the same condition, the averaged approximation error of leveraging a preconditioned variance estimate is decreasing over time.

\begin{theorem}
\label{thm:auto_switch_motivation}
   Suppose $\*g_t^2$ is stationary and has bounded norm $\|\*g_t^2\|_\infty\leq G$ for some constant $G > 0$. Given a sufficient precondition step $t_0$ such that $t_0>\log_{\beta_2}\left(1-\frac{1}{\sqrt{2}}\right)$, then for any step $t>t_0$ it holds with probability at least $1-\delta$,
    \begin{align*}
        \|\hat{\*v}_t - \hat{\*v}_{t_0}\|_\infty < \sqrt{4G^2(1-\beta_2)^2(t-t_0)\log\left(\frac{2}{\delta}\right)}.
    \end{align*}
\end{theorem}

Theorem~\ref{thm:auto_switch_motivation} provides the worst-case accumulated error of using preconditioned $\*v_{t_0}$ to estimate $\*v_t$ ($\forall t>t_0$). 
Observing the bound given in Theorem~\ref{thm:auto_switch_motivation}, conditioned on $t_0$, the maximal accumulated change to a variance coordinate is sublinear to time $t-t_0$.
This suggests when we use $\*v_{t_0}$ to estimate $\*v_t$ for any $t>0$, the average error obtained in each step is decreasing over time with rate $O(1/\sqrt{t-t_0})$.

On the other hand, the coefficient $(1-\beta_2)^2$ is a very small number both theoretically and empirically. In theory, it is provably shown that to ensure Adam convergence, $1-\beta_2$ has to be small enough such that $1-\beta_2 = O(N^{-3})$, where $N$ is the size of the training dataset \citep{zhang2022adam}, and having a larger $1-\beta_2$ could lead to divergence.
In practice $\beta_2$ is often set to a value such that $(1-\beta_2)^2$ reduces $t-t_0$ by orders of magnitude: For instance, the default setting of $\beta_2$ is $0.999$ given in the original Adam paper \citep{kingma2014adam} and most of the deep learning libraries \citep{paszke2019pytorch,flax2020github}, leading to $(1-\beta_2)^2=10^{-6}$; on foundation models like GPT-3 and Megatron, $(1-\beta_2)^2$ is around $10^{-4}$ \citep{brown2020language,smith2022using}.

Building upon this, the overall structure of {\step} algorithm is shown in Algorithm~\ref{algo:STEP} that separates the training into two phases. In the first phase (the first while loop), the normal Adam is used and the variance estimate is actively updated; in the second phase (the second while loop), the variance estimate obtained from phase I is then used as a precondition to learn the mask with Straight Through Estimator (STE).

\section{Auto Switch Between two Phases}
\label{sec:as}
\begin{algorithm}[t!]
	\caption{Proposed {\as} subroutine for {\step}}\label{algo:AutoSwitch}
	\begin{algorithmic}[1]
		\Require Sample size $T_w=\lfloor(1-\beta_2)^{-1}\rfloor$ given by {\step}, the current step $t$, (Optional: lower bound $T_{\min}$ and upper bound $T_{\max}$ for clipping).
	    \State Compute the current sample on the variance change: $$\textbf{Option I: }Z_t = d^{-1}\|\*v_t - \*v_{t-1}\|_1;$$ $$\textbf{Option II: }Z_t = \exp(d^{-1}\|\log(\*v_t - \*v_{t-1})\|_1).$$
	    \State Estimate mean over the sliding window: $$\bar{Z} = T_w^{-1}\sum\nolimits_{j=t-T_w+1}^{t}Z_j.$$
	    \If{(Optional) Use Clipping}
	        \State \textbf{return} $t > T_{\max}$ \textbf{or} $\bar{Z} < \epsilon$ \textbf{and} $t > T_{\min}$.
	    \Else
            \State \textbf{return} $\bar{Z} < \epsilon$.
        \EndIf
	\end{algorithmic}
\end{algorithm}

In the previous section, we've discussed the theoretical motivation of using preconditioned variance on learning N:M masks with Adam. However, the central question is still left open: how should we set the switching point $t_0$ in Theorem~\ref{thm:auto_switch_motivation}.
As partially discussed in Section~\ref{sec:introduction}, while identifying reliable Adam variance during training is an established problem, 
most of the existing methods solve this via heuristics or hyperparameter tuning \citep{tang20211,li20211,lu2022maximizing}.
In this section, we introduce {\as}, a subroutine that automatically decides the switching point between precondition and mask learning phases by testing the variance change concentration along the training trajectory.

\begin{figure}[t!]
  \centering
  \subfigure[ResNet18 on CIFAR10]{
    \includegraphics[width=0.46\textwidth]{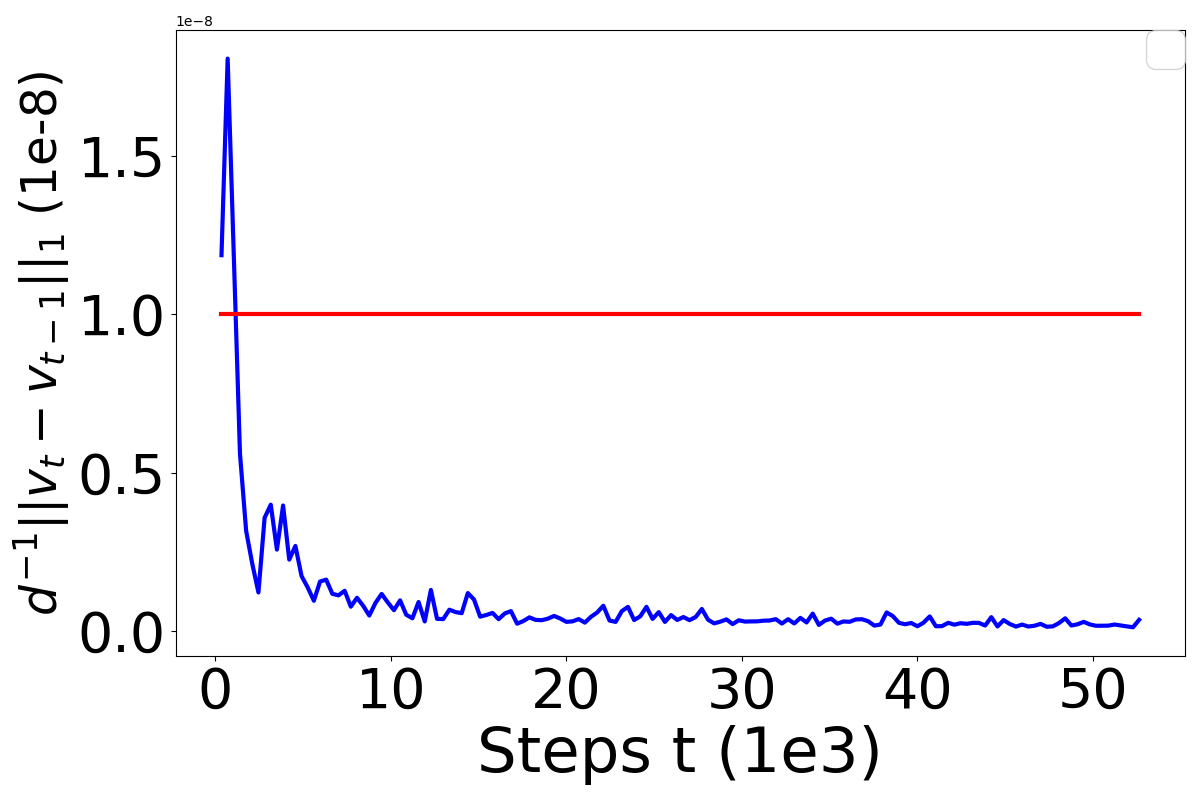}
  }
  \subfigure[DenseNet121 on CIFAR100]{
    \includegraphics[width=0.46\textwidth]{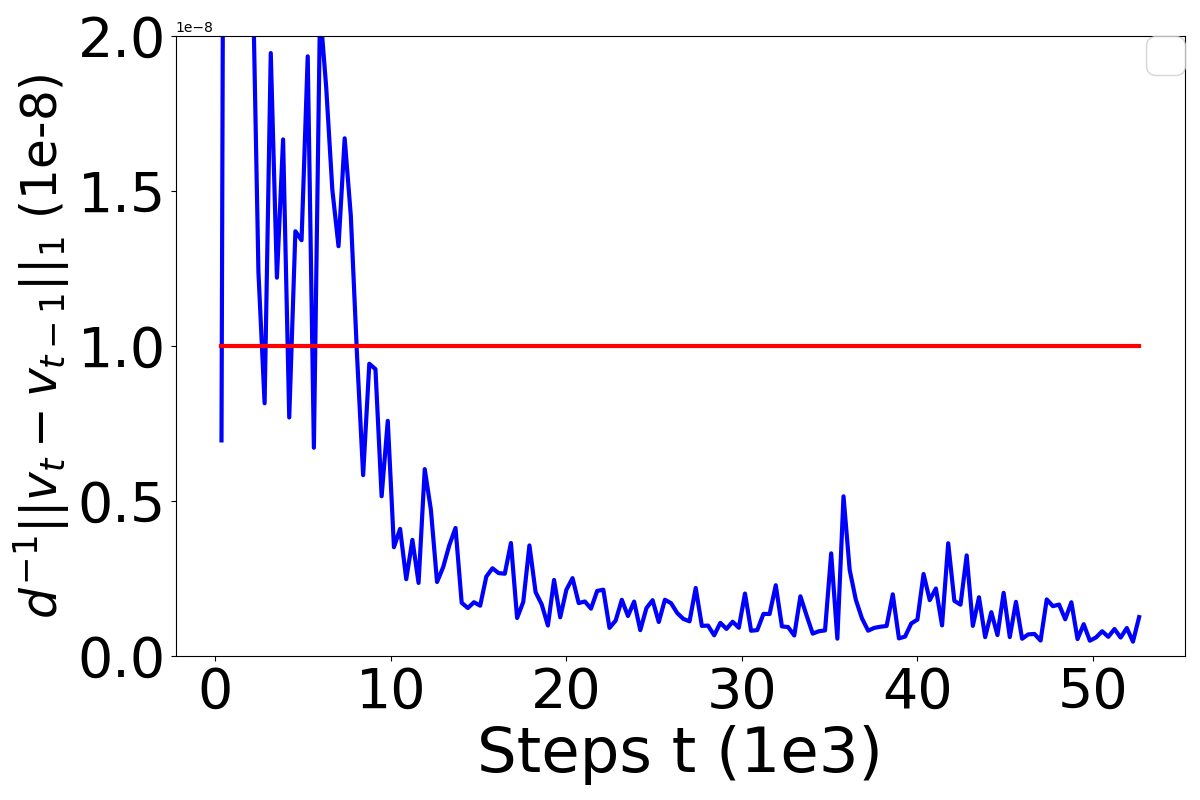}
  }
  \caption{Figure showing per-coordinated variance difference  $d^{-1}\|\*v_t-\*v_{t-1}\|_1$ over steps (in blue curves), in the CIFAR tasks shown in Figure~\ref{fig:srste_issue_demo}. We also plot the $\epsilon$ (in the red line). We observe the update to each coordinate of the variance is quickly dominated by the $\epsilon$.}
  \label{fig:var_diff}
\end{figure}

\paragraph{Baseline Methods and Their Limitations.}
We start with the methods in the literature on identifying the switching point.
A straightforward way to do this is leveraging standard hyperparameter tuning protocol such as grid search or random search \citep{bergstra2012random}: setting a few candidate steps and iterate over them and choose the one yielding best performance.
However, adding hyperparameters heavily relies on heuristics and requires certain domain knowledge for practitioners. 

There have been a few efforts on identifying a good switching point by monitoring the variance metrics.
The first is to monitor the relative error as proposed in \citep{agarwal2021adaptive}, which identifies step $t$ as the end of the critical regime if:
\begin{align}
\label{equ:switch:baseline:relative}
    \frac{|\|\*v_t\| - \|\*v_{t-1}\||}{\|\*v_{t-1}\|} < 0.5,
\end{align}
where the bound $0.5$ given by \citep{agarwal2021adaptive}. 
The intuition is to use the tensor norm difference to approximate the tensor difference (note that storing $\*v_t$ and $\*v_{t-1}$ directly could incur non-trival memory overhead due to the high-dimensionality).
Another similar method is proposed in \citep{tang20211}, which suggests a staleness comparison on the variance norm. Concretely, \citet{tang20211} identifies step $t$ as the end of the critical regime if:
\begin{align}
\label{equ:switch:baseline:stale}
    \frac{\left\|\*v_t \right\|_1}{\left\|\*v_{t-\lfloor (1-\beta_2)^{-1}\rfloor}\right\|_1} > 0.96,
\end{align}
where the criteria 0.96 is provided by \citep{tang20211}.

The baseline methods (Equation~(\ref{equ:switch:baseline:relative}) and (\ref{equ:switch:baseline:stale})) are limited in practice in three-fold: (i) when evaluating the switching point $t$, it can be easily affected by the noise at step $t$; (ii) Although both of the methods require relative metrics, the thresholds are still hand-picked, and thus introducing additional noise to the criterion; (iii) Both of the methods use the tensor norm over all the coordinates. On one hand, norm can be a good indicator for status of variance but not for variance changes. On the other hand, the switching point can easily mistakenly be missed due to the outliers among the coordinates, especially on large models, where the order of variance magnitude varies significantly \citep{xiong2020layer,liu2020understanding}.

\paragraph{AutoSwitch.}
The main procedures of {\as} are summarized in Algorithm~\ref{algo:AutoSwitch}. To cope with the gradient noise and outlier coordinates, {\as} samples over time $t$ the per-coordinate variance change via arithmetic mean (\textbf{Option I}) or geometric mean (\textbf{Option II}). While geometric mean is robust to outliers, in practice we found arithmetic mean is sufficient for deciding the switching point. We set the sampling window length to be $\lfloor (1-\beta_2)^{-1}\rfloor$. This quantity is motivated from the Markov Chain theory: if we model the dynamic of $\*v_t$ as a Markov Chain, then the mixing time of the chain then is roughly $\tilde{O}\left(\frac{1}{1-\beta_2}\right)$.

While sampling mitigates the noise from single step evaluation, it still remains unclear what metric we should be applying to decide the phase length. Note that in the baseline works (Equation~(\ref{equ:switch:baseline:relative}) and (\ref{equ:switch:baseline:stale})), hand-picking values are applied.
Ideally, we should leverage some metrics from the Adam optimizer that is adapted to each task.
Based on this, {\as} uses the $\epsilon$ from Adam as the signal. The $\epsilon$ is originally used in Adam to prevent zero division. In some research it has been found that it largely decides the model convergence. To justify our motivation, we plot the per-coordinate variance change and $\epsilon$ in Figure~\ref{fig:var_diff}.
We observe the update to each coordinate of the variance is quickly dominated by the $\epsilon$ as the training proceeds.

\paragraph{Clipping for Tight Training Budget.}
While Algorithm~\ref{algo:AutoSwitch} provides a statistical way of identifying the switching point, in practice, varying training budgets (e.g. model fine-tuning) are usually considered. We can use clipping to clamp a computed switching point $t_0$ between given $T_{\min}$ and $T_{\max}$.
The clipping bounds are two optional variables that regularize the {\as} subroutine.
By default, we suggest using $T_{\min}=0.1T$ and $T_{\max}=0.5T$, these two values are motivated by Geweke's convergence diagnostic in MCMC theory \citep{geweke1991evaluating}. Recall that the update of $\*v_t$ forms a markov chain, and so concentration of the first 10\% and last 50\% of the chain can be used as a good indicator on the convergence \citep{geweke1991evaluating}.
\begin{figure}[t!]
  \centering
  \subfigure[ResNet18 on CIFAR10]{
    \includegraphics[width=0.46\textwidth]{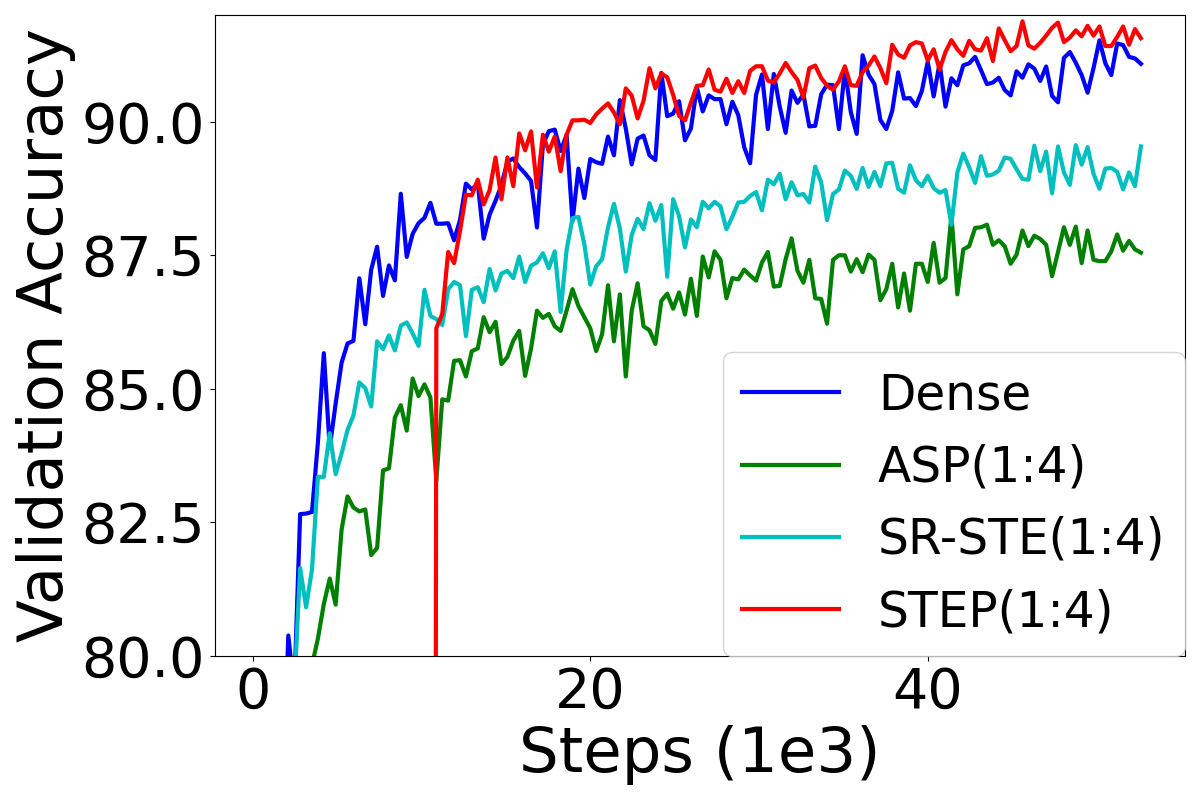}
  }
  \subfigure[DenseNet121 on CIFAR100]{
    \includegraphics[width=0.46\textwidth]{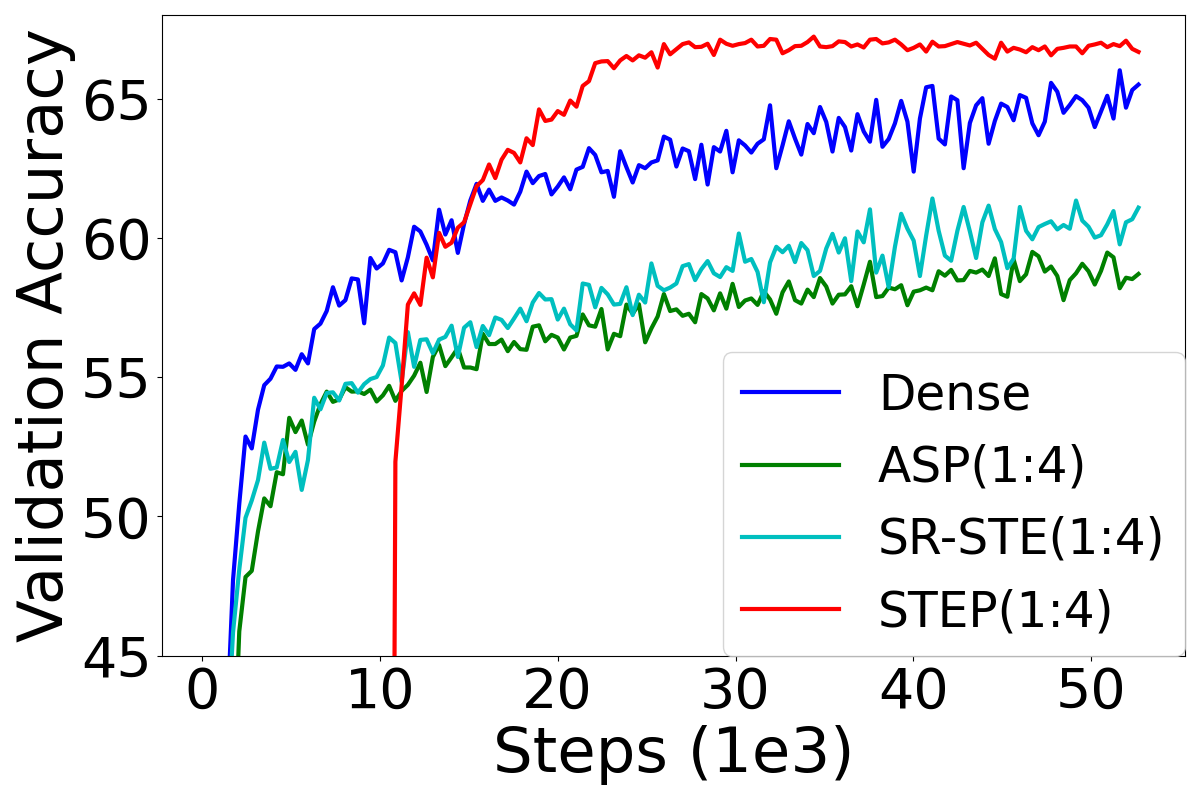}
  }
  \caption{Figure showing how {\step} mitigates the gap of baseline algorithm ASP \citep{mishra2021accelerating} and SR-STE \citep{zhou2021learning}. In this experiment, 1:4 sparsity is used. The switching point of {\step} is decided by the {\as} subroutine. Note that during the precondition phase of {\step}, the model does not involve the mask learning while the model is evaluated with sparsity (for fair comparison to baseline models). And thus the evaluation accuracy during that phase is low compared to the mask learning phase.}
  \label{fig:exp:res_dense_net}
\end{figure}

\begin{figure}[t!]
  \centering
  \subfigure[ResNet18 on CIFAR10]{
    \includegraphics[width=0.46\textwidth]{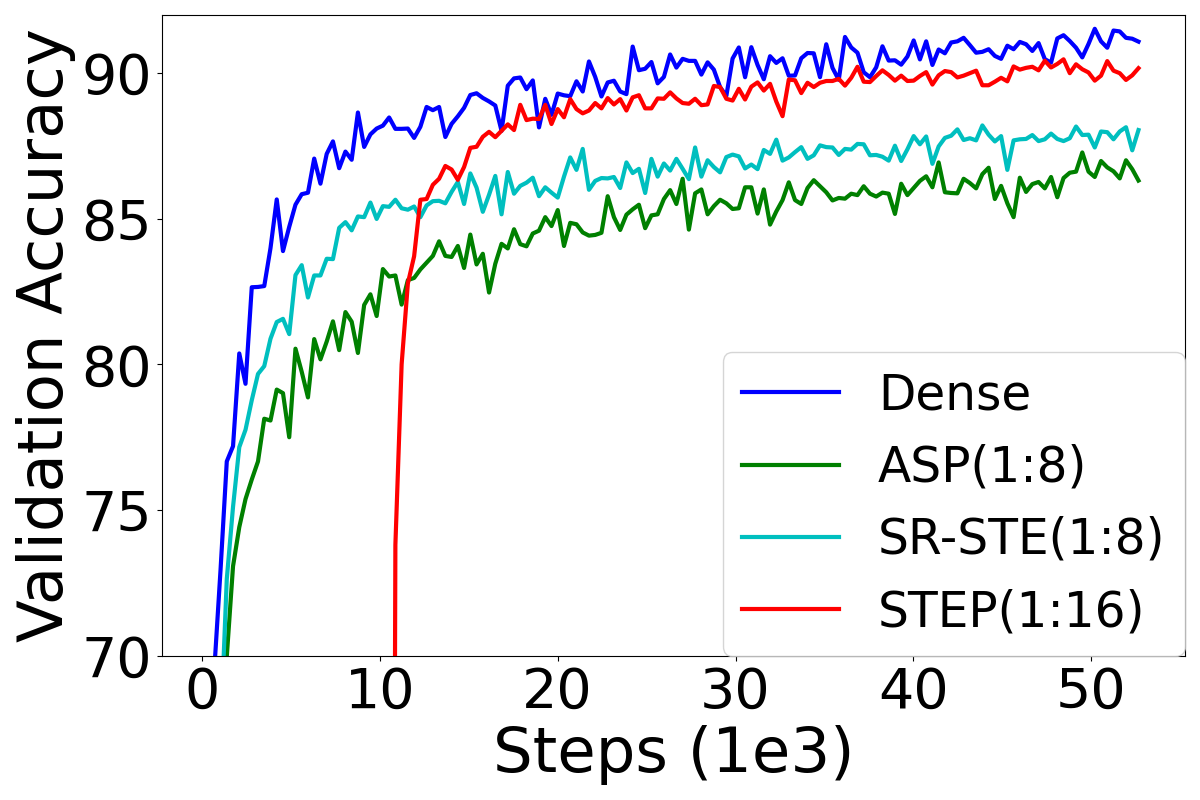}
  }
  \subfigure[DenseNet121 on CIFAR100]{
    \includegraphics[width=0.46\textwidth]{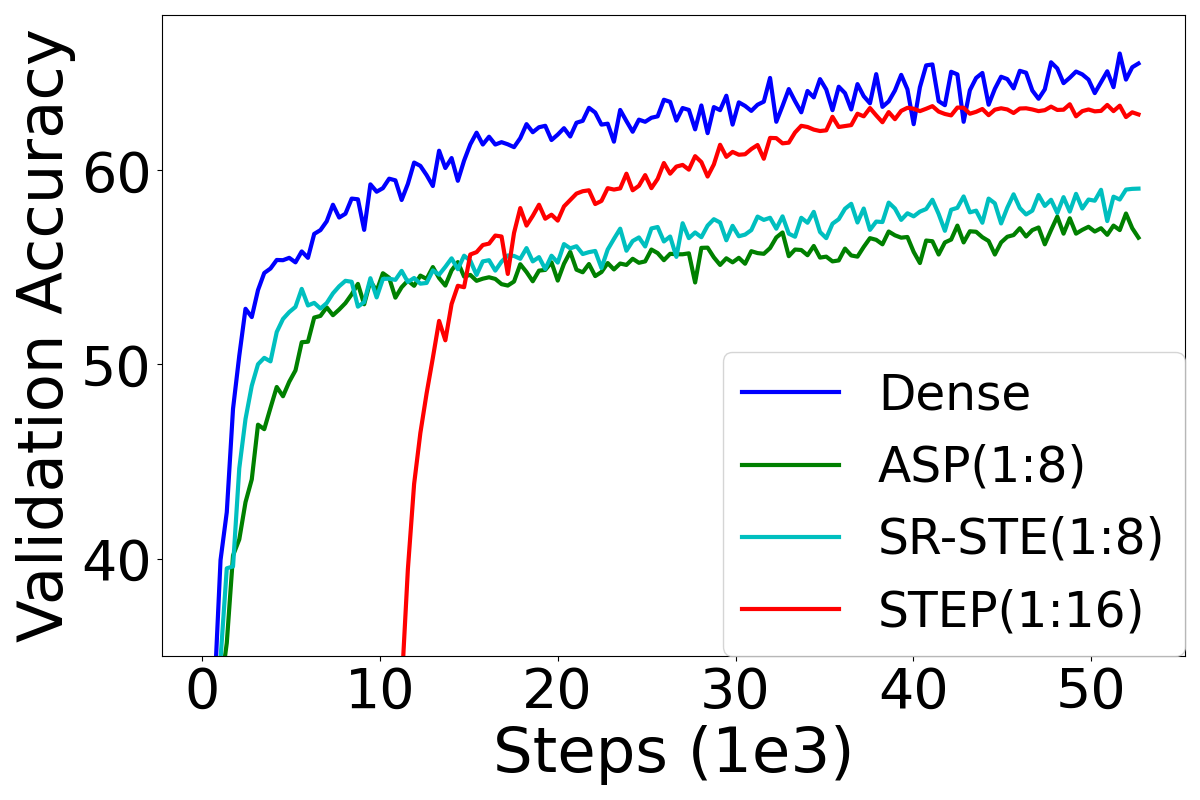}
  }
  \caption{Figure comparing the performance of {\step} under aggressive sparsity ratio. Comparing the results with Figure~\ref{fig:exp:res_dense_net}, it suggests the {\step} recipe is robust to aggressive sparsity ratio up to 1:16, while baselines degrade the evaluation accuracy at 1:8.}
  \label{fig:exp:aggressive_ratio}
\end{figure}

\begin{table}[t]
  \centering
  \caption{Comparing {\as} (Algorithm~\ref{algo:AutoSwitch}) with two baseline approaches Equation~(\ref{equ:switch:baseline:relative}) \citep{agarwal2021adaptive} and (\ref{equ:switch:baseline:stale}) \citep{tang20211}. We measure the average change within 1k steps after the precondition $t_0$ identified by different approaches: $10^{-3}\sum_{t=t_0}^{t_0+1000}\|\*v_{t+1} - \*v_t\|_1$. A lower number indicates better estimation for the switching points. The numbers for each experiment are averaged over 5 different random seeds.}
  \label{exp:table:autoswitch}
  \begin{tabular}{lccccccccccc}
  \hline  
  Task & Eq.~(\ref{equ:switch:baseline:relative}) & Eq.~(\ref{equ:switch:baseline:stale}) & \textbf{AS}\\
  \hline  
ResNet18/CF10 & 1.58e-1 & 5.58e-2 & \textbf{0.79e-2} \\
DenseNet121/CF100 & 5.26e-1 & 1.28e-2 & \textbf{0.46e-2} \\
BERT-Large (PreT) & 4.92e-6 & 2.71e-7 & \textbf{2.28e-7} \\
  \hline
  \end{tabular}
\end{table}

\begin{table*}[t]
  \centering
  \caption{Finetuning BERT-Base on the GLUE development set. The original results are from \citep{devlin2018bert}. The Dense results are reproduced by ours with no sparsity. For different recipes (ASP, SR-STE and {\step}), 2:4 sparsity is applied on all the linear modules (including attention, intermediate and output layer of BERT.) The scores are the median scores over 10 runs with different seeds. We observe compared to baselines, {\step} has a negligible drop on the average score compared to the dense counterpart.}
  \label{exp:table:glue}
  \begin{tabular}{lccccccccccc}
  \hline  
  & RTE & MRPC & STS-B & CoLA & SST-2 & QNLI & QQP & MNLI-m & MNLI-mm & Avg Score\\
  \hline  
Original & 66.4 & 84.8 & 85.8 & 52.1 & 93.5 & 90.5 & 89.2 & 84.6 & 83.4 & 81.1 \\
Dense & 65.0 & 85.1 & 85.2 & 51.0 & 92.3 & 91.1 & 91.0 & 84.6 & 83.6 & 81.0 \\
  \hline
ASP & 57.4 & 79.2 & 81.7 & 47.2 & 88.5 & 83.7 & 84.8 & 80.6 & 79.5 & 75.8 \\
SR-STE & 55.6 & 81.3 & 88.2 & 47.8 & 90.2 & 86.6 & 90.1 & 82.1 & 82.9 & 78.3 \\
{\step} & 62.4 & 84.7 & 88.7 & 50.4 & 91.8 & 89.2 & 90.9 & 84.2 & 83.9 & \textbf{80.7} \\
  \hline
  \end{tabular}
\end{table*}

\begin{table}[t]
  \centering
  \caption{Training different language modeling tasks on Wikitext-2(-103). For different recipes (ASP, SR-STE and {\step}), 2:4 sparsity is applied on all the \texttt{Conv1D} modules of GPT2. The numbers are averaged evaluation perplexity over 10 runs with different seeds. }
  \label{exp:table:wikitext}
  \begin{tabular}{lccccccccccc}
  \hline  
  & Wikitext-2 & Wikitext-103 \\
  \hline  
Dense & 21.15 & 16.57 \\
\hline
ASP & 37.09 & 26.29  \\
SR-STE & 28.54 & 18.93  \\
{\step} & \textbf{23.85} & \textbf{17.02} \\
  \hline
  \end{tabular}
\end{table}

\begin{table}[t]
  \centering
  \caption{Extension of {\step} to layer-wise N:M masks learning. The N:M sparsity ratios are decided in a per-layer fashion following the strategy given in \citep{sun2021dominosearch}. The numbers in this table are averaged over 5 runs. The results suggest {\step} can provide in-place improvement when combined with per-layer structured sparsity.}
  \label{exp:table:domino}
  \begin{tabular}{lccccccccccc}
  \hline  
  & N:M & RN-CF10 & DN-CF100 \\
  \hline  
Dense & / & 91.56 & 65.62 \\
\hline
DS & Mixed N:8 & 89.94 & 64.88  \\
DS+{\step} & Mixed N:8 & \textbf{91.42} & \textbf{65.71} \\
\hline
DS & Mixed N:16 & 87.08 & 62.13  \\
DS+{\step} & Mixed N:16 & \textbf{90.93} & \textbf{65.04} \\
\hline
DS & Mixed N:32 & 85.37 & 60.47  \\
DS+{\step} & Mixed N:32 & \textbf{90.12} & \textbf{64.91} \\
\hline
  \end{tabular}
\end{table}

\section{Experiment}
\label{sec:experiment}
In this section we evaluate the effectiveness of proposed {\step} and {\as} on various tasks, comparing it to other baseline recipes of learning N:M masks. We also show that {\step} can be easily extended to incorporate other techniques such as layer-wise sparsity \citep{sun2021dominosearch}.
All of the experiments run on a Google Cloud TPUv3-8 virtual machine.

\paragraph{Overview of Tasks.}
Throughout these sections, we adopt the following tasks for the evaluation: (1) Training various vision models (ResNet18, Densenet121) on CIFAR10/100 dataset \citep{krizhevsky2009learning}. (2) Finetuning BERT-Base\citep{devlin2018bert} on the GLUE benchmark \citep{wang2018glue}. (3) Training a 6-layer Transformer model on the WMT17 De-En Translation task following \citep{vaswani2017attention}. (4) Finetuning GPT-2 model \citep{radford2019language} on Wikitext-2 and Wikitext-103 \citep{merity2016pointer}.

\paragraph{Hyperparameters.}
We apply the grid search over the following hyperparameters on each task. Notice that we only tune the hyperparameters for the baselines, but not for {\step}. That is, {\step} reuses the hyperparameters tuned for SR-STE. This suggests {\step} can provide in-place improvement over the baseline recipes. For all the Adam-specific hyperparameters we adopt the default values: \{$\beta_1=0.9$, $\beta_2=0.999$, $\epsilon=1e-8$\}. 
For the CIFAR tasks, we adopted batch size 128 and tune the learning rate from \{$1e-4$, $5e-5$, $1e-5$\}; for BERT and GPT-2 fine-tuning we follow \citep{tang20211} and tune batch size from \{$8,16,32$\} and learning rate from \{$1e-4,5e-5,1e-5$\}; for WMT machine translation we follow the exact setup\footnote{A more detailed description can be found in Section 4 \citep{kao2022training}.} of \citep{vaswani2017attention} and \citep{kao2022training}.

\paragraph{The Effectiveness of AutoSwitch.}
We start from evaluating the effectiveness of {\as} over baseline methods as introduced in Section~\ref{sec:as}.
Concretely, we compare Algorithm~\ref{algo:AutoSwitch} with Equation~(\ref{equ:switch:baseline:relative}) proposed by \citep{agarwal2021adaptive} and Equation~(\ref{equ:switch:baseline:stale}) proposed by \citep{tang20211}. For each task, we first profile the $\|\*v_t\|_2$, $\|\*v_t\|_1$ and $\|\*v_{t+1} - \*v_t\|_1$ for all the $t\geq 1$ since these suffice for running the three approaches. Then for any $t_0$ as a precondition step found by each method, we compute the average variance change in the next 1k steps, i.e., $10^{-3}\sum_{t=t_0}^{t_0+1000}\|\*v_{t+1} - \*v_t\|_1$ as measuring the reliability of preconditioned variance. Intuitively, a smaller average variance change implies better preconditioning.
We summarize the results in Table~\ref{exp:table:autoswitch}, the results suggest {\as} is able to identify variance with subtle changes in the following steps compared to the other two baselines.

\begin{figure}[t!]
  \centering
  \subfigure[N:M=1:32]{
    \includegraphics[width=0.46\textwidth]{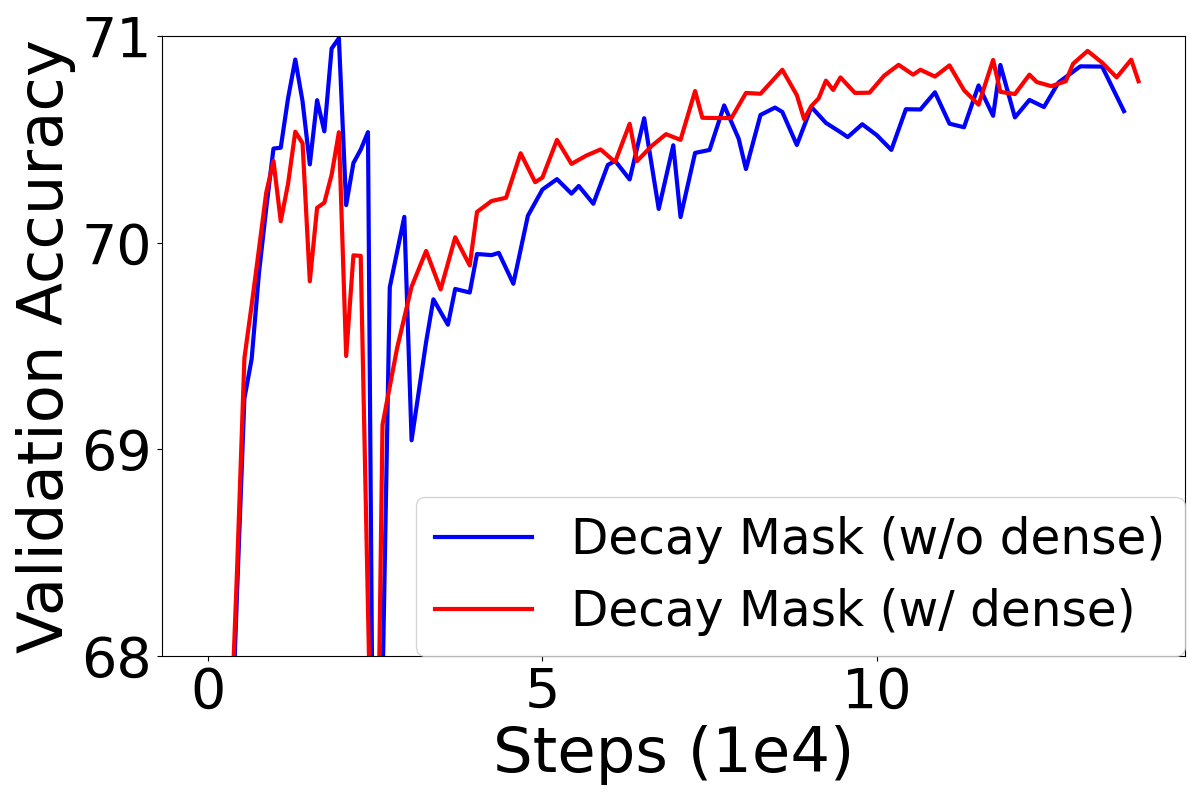}
  }
  \subfigure[N:M=1:64]{
    \includegraphics[width=0.46\textwidth]{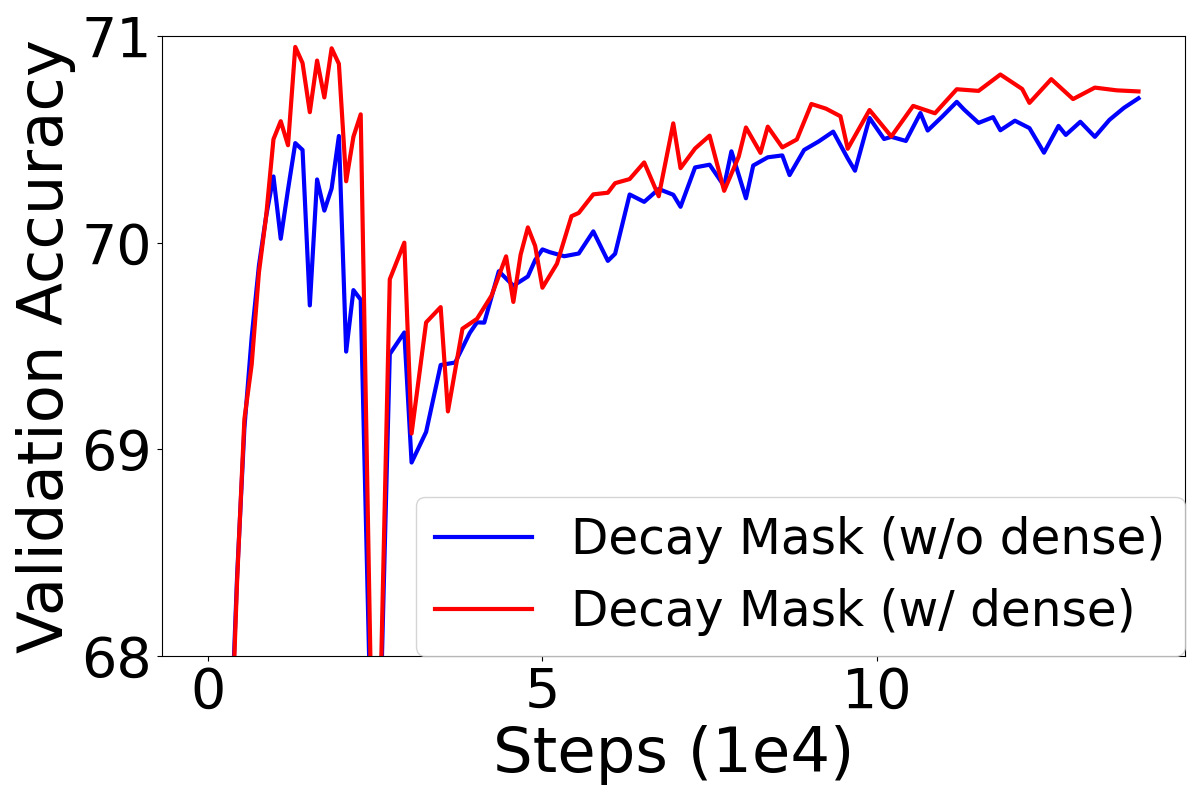}
  }
  \caption{Ablation Study on Decaying Mask. We follow the setting of \citep{kao2022training} and train the 6-layer Transformer model on the WMT17 De-En translation task. To shows the importance of preconditioning with dense updates. We include the results and compare the Decaying Mask recipe with and without the dense training phase.}
  \label{fig:exp:ablation:decaying_mask}
\end{figure}

\paragraph{Comparing with Baselines.}
We now evaluate the performance of {\step} with the following baseline recipes: Dense (no mask is learnt), ASP \citep{mishra2021accelerating} and SR-STE \citep{zhou2021learning}. The comparison is carried out on three tasks: training ResNet18 and Densenet121 from scratch on CIFAR10/100; finetuning BERT-Base on GLUE; and finetuning GPT2 on Wikitext-2/-103. For all the recipes, we apply 2:4 sparsity \cite{pool2020accelerating} to all the modules. 
More concretely: for ResNet and DenseNet, the sparsity is applied on all the \texttt{Conv2D} layers; for BERT-Base, all the \texttt{Linear} modules in attention, intermediate and output layers are sparsified; in GPT-2, the sparsity is applied on all the \texttt{Conv1D} modules.
We summarize the results in Figure~\ref{fig:exp:res_dense_net}, Table~\ref{exp:table:glue} and ~\ref{exp:table:wikitext}. The results consistently suggest under the same sparsity ratio, {\step} is able to mitigate the accuracy drop between baseline recipes (ASP and SR-STE) and dense training.
Perhaps surprisingly, we found in the DenseNet task, {\step} achieves higher validation accuracy compared to the dense training.

\paragraph{Robustness to Aggressive Structured Pruning.}
We extend the previous experiments on pre-training ResNet18 and DenseNet121 with different sparsity ratios, using {\step} recipes. We summarize the results in Figure~\ref{fig:exp:res_dense_net}, we observe up to N:M=1:16, {\step} recipe has negligible accuracy drop compared to the dense training, while other recipes have non-trivial evaluation accuracy gap at 1:8.

\paragraph{Ablation Study I: Layer-wise Pruning.}
We now demonstrate that {\step} can be trivially extended to layer-wise SR-STE as considered in DominoSearch \citep{sun2021dominosearch}. We now run the {\step} and {\as} following a per-module fashion, with per-layer sparsity ratio determined by the DominoSearch algorithm \citep{sun2021dominosearch}. We summarize the results of using plain DominoSearch (DS) and DS combined with {\step} in Table~\ref{exp:table:domino}. The results there suggest combined with {\step}, DominoSearch can have more stable results, especially over aggressive N:M ratios. 
More concretely, when the sparsity ratios are increased to N:32, the original DominoSearch already incurs over 5\% accuracy drop while with {\step}, the accuracy drop is generally around 1\% on both ResNet and DenseNet.
Notice that {\step} does not modify the dynamic sparsity ratio assignment strategy as used in the original DominoSearch. 
This, on the other hand, implies {\step} provides in-place improvement over layer-wise sparsity.

\paragraph{Ablation Study II: Decaying Mask.}
In this experiment, we conduct an ablation study on a recently proposed recipe named Decaying Mask \citep{kao2022training}. The recipe proceeds as follows: first run dense training for some iterations, and then start the sparse training phase.
At the beginning of the sparse training phase, it starts with M-1:M structured sparsity. As training progresses, Decaying Mask increases
the sparsification degree by applying N:M structured sparsity at different decaying intervals, where $N=\left\lfloor \frac{M}{2^s} \right\rfloor$.

Note that the original Decaying Mask recipe already includes the dense training phase. In this ablation study, we follow the setup of \citep{kao2022training} and compare how Decaying Mask behaves with and without its dense training phase. We summarize the results in Figure~\ref{fig:exp:ablation:decaying_mask}. It suggests if no dense training is performed at the beginning of the recipe, there will be a certain accuracy drop even if the sparsity ratio is gradually decreased. This, again, substantiates the motivation of {\step} recipe.

\paragraph{Ablation Study III: Varying Preconditioning Phase Length.}

\begin{figure}[t!]
  \centering
  \subfigure[ResNet18 on CIFAR10]{
    \includegraphics[width=0.46\textwidth]{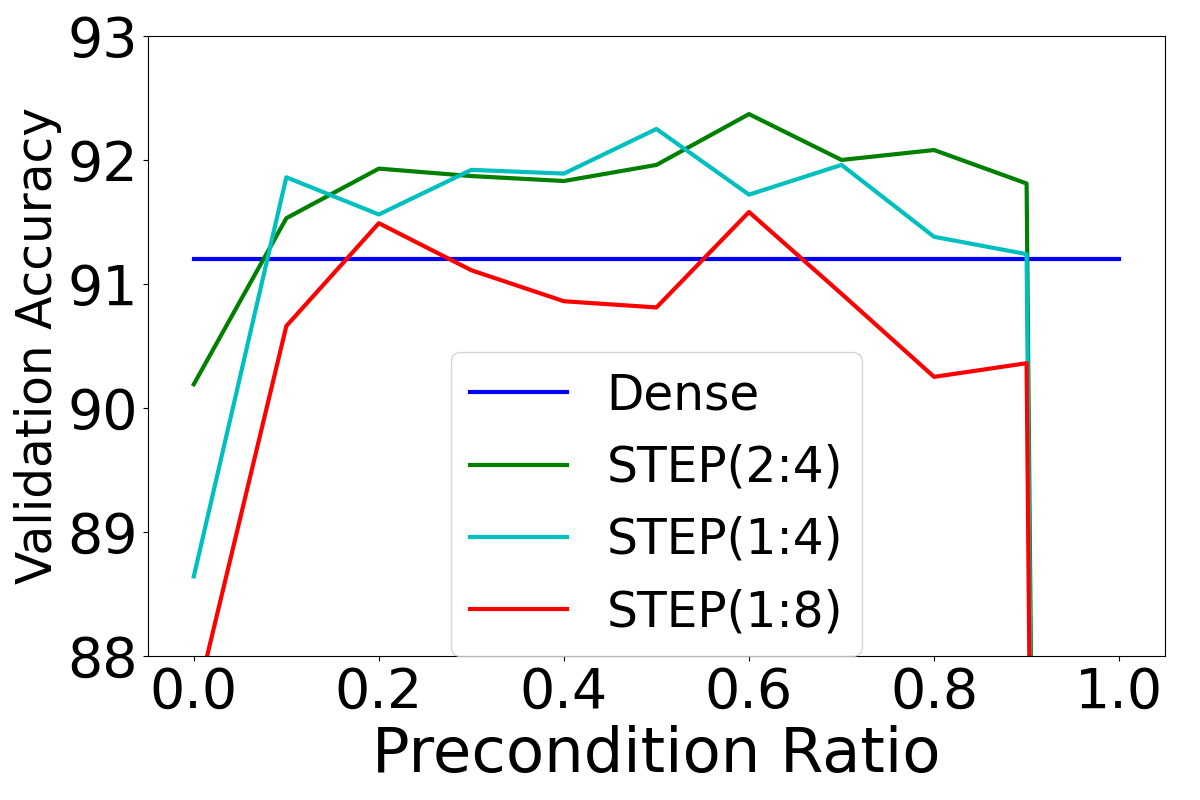}
  }
  \subfigure[DenseNet121 on CIFAR100]{
    \includegraphics[width=0.46\textwidth]{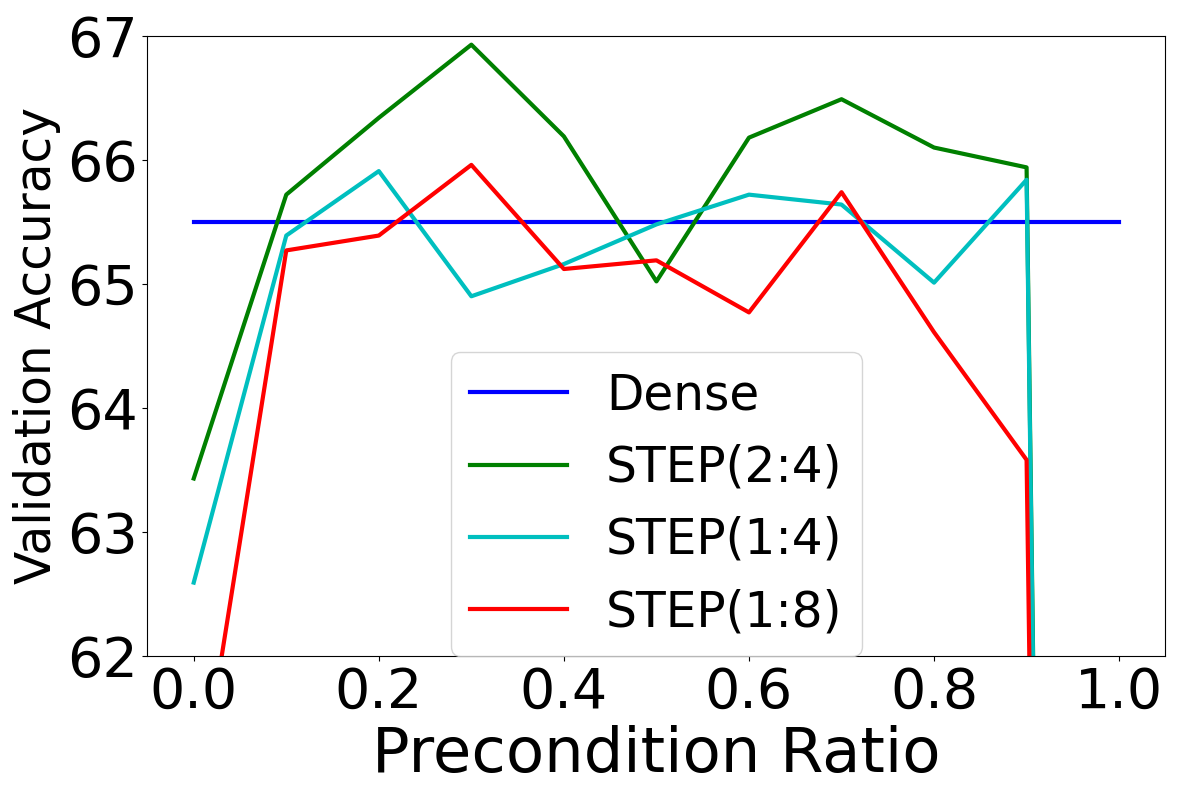}
  }
  \caption{Ablation study on different precondition phase length. The X-axis denotes the ratio of precondition phase length over the total number of training steps; while the Y-axis denotes the evaluation accuracy of the output model at the end. We observe that the switching point between precondition and mask learning phase is quite flexible.}
  \label{fig:exp:ablation:precondition_ratio}
\end{figure}

We continue investigating the effect of preconditioning phase length on the final model accuracy.
We repeat the CIFAR experiments on two vision models and rerun the {\step} algorithm with different precondition phase length. We summarize the results in Figure~\ref{fig:exp:ablation:precondition_ratio}.
We observe that {\step} is able to achieve dense accuracy when the ratio of preconditioning phase is between 10\% and 80\% (despite the fact that {\as} decides the ending point to be around 20\%).
This suggests the switching point in {\step} is quite flexible over the entire training trajectory, and is robust to the potential noise in the {\as} subroutine.

\paragraph{Ablation Study IV: Why Fixing the Variance.}

\begin{figure}[t!]
  \centering
  \subfigure[ResNet18 on CIFAR10]{
    \includegraphics[width=0.46\textwidth]{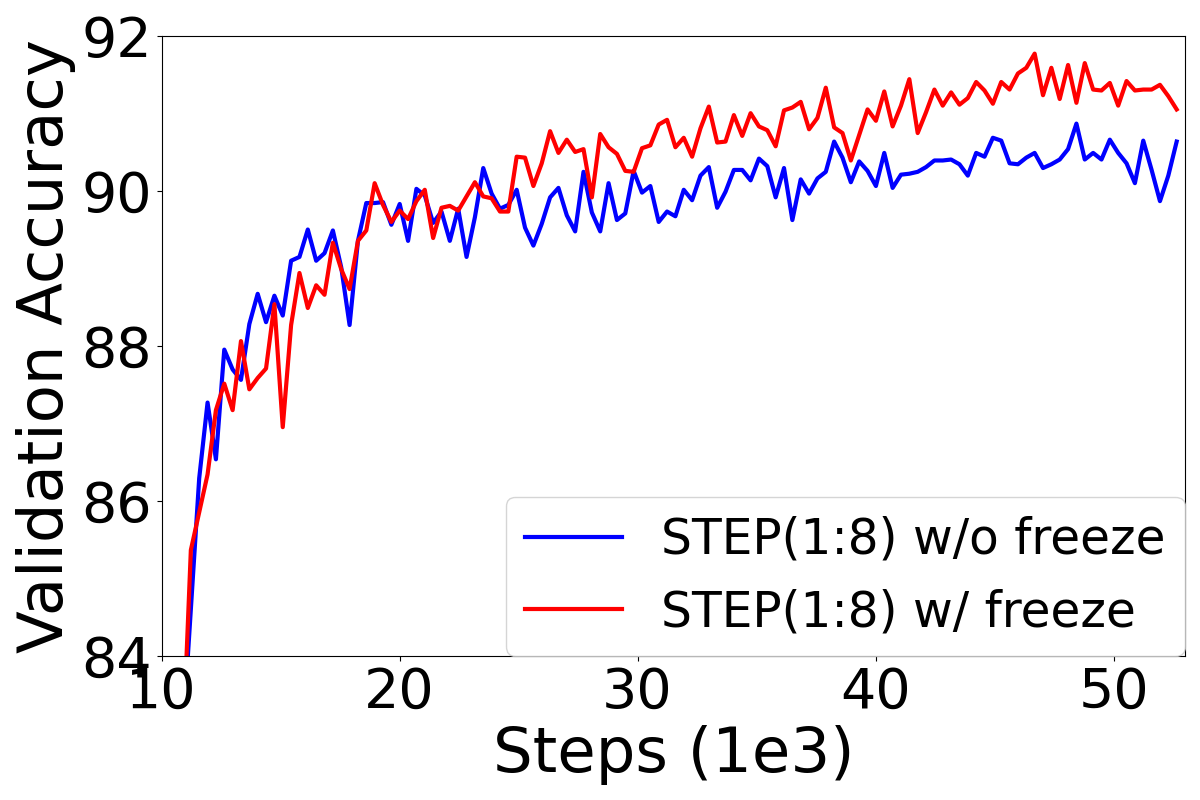}
  }
  \subfigure[DenseNet121 on CIFAR100]{
    \includegraphics[width=0.46\textwidth]{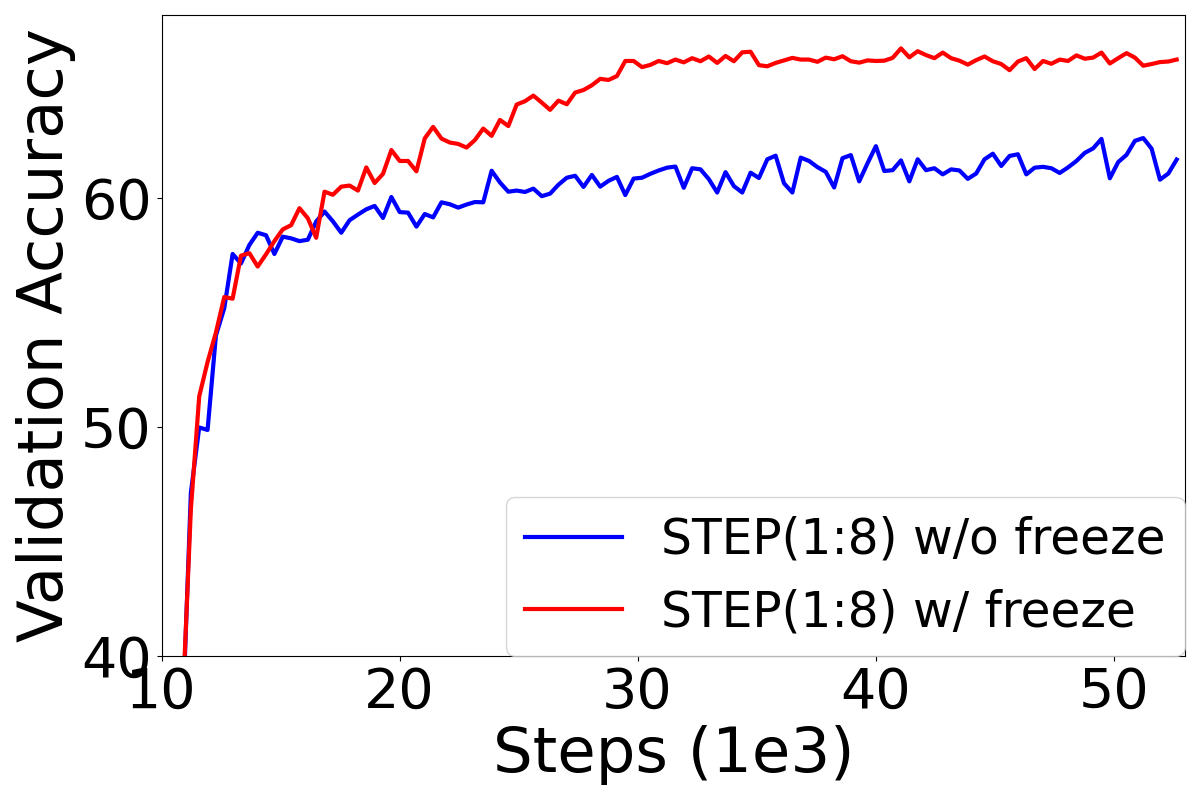}
  }
  \caption{Ablation study on comparing with and without updating variance term during the mask learning phase. The curves suggest freezing (fixing) the preconditioned variance during the mask learning phase is crucial.}
  \label{fig:exp:ablation:variance_fix}
\end{figure}

Note that in the original {\step} Algorithm, the variance remains fixed during the masking learning phase. A natural question to this would be: does it help if we keep updating the variance using the gradients computed on the sparsified model? In practice, we observe this in fact has negative impact. We rerun the ResNet/DenseNet experiments with two variants: original {\step} and {\step} where variance is updated in the second phase. We summarize the results in Figure~\ref{fig:exp:ablation:variance_fix}.
It suggests keeping updating the variance with gradients computed on masked weights reduces the final evaluation accuracy, which implies the noise level in gradients remains high during mask learning, even in the later stage of training.
\section{Conclusion}
In this paper, we identify the state-of-the-art recipe SR-STE incurs non-trivial model degradation when applied in Adam-based model training. We propose an algorithm named {\step} that separates the training into two phases, where in the first phase, the Adam optimizer precondition a reliable second moment (variance) estimate; while in the second phase, such variance remains fixed and is used as a precondition to learn the N:M structured sparsity masks. We also propose a subroutine named {\as} that automatically determines the switching point of two phases. Compared to other approaches, {\as} shows stable and reliable estimation. Empirically we evaluate {\step} on various benchmarks including text classification, image classification and language modeling. We demonstrate {\step} mitigates the accuracy drop compared to other recipes and is robust to aggressive sparsity ratios. We also show that {\step} can be easily integrated with other techniques such as layer-wise sparsity.

\bibliography{refs}

\newpage
\appendix
\onecolumn
\section{Technical Proof}
\subsection{Proof to Theorem~\ref{thm:auto_switch_motivation}}

\begin{proof}
We first define the filtration $\mathcal{F}_t$ over step $t\in\{1, \cdots, T\}$, where the randomness come from the sampling of the data point $\zeta_t$. And next we show the update for each coordinate of $\*v_t$ is a martingale difference sequence.
From the update of Adam, we get:
\begin{align*}
    \hat{\*v}_{t+1} - \hat{\*v}_{t} = & \frac{\*v_{t+1}}{1-\beta_2^{t+1}} - \frac{\*v_{t}}{1-\beta_2^{t}} \\
= & \frac{1}{1-\beta_2^{t+1}} \left( \*v_{t+1} - \frac{1-\beta_2^{t+1}}{1-\beta_2^t}\*v_t \right) \\
= & \frac{1}{1-\beta_2^{t+1}} \left[ \beta_2\*v_{t} + (1-\beta_2)\*g_t^2 - \frac{1-\beta_2^{t+1}}{1-\beta_2^t}\*v_t \right] \\
    = & \frac{1}{1-\beta_2^{t+1}} \left[ (1-\beta_2) \cdot \left(\*g_t^2 - \frac{\*v_t}{1-\beta_2^t}  \right) \right].
\end{align*}
Take expectation with respect to the filtration, we obtain
\begin{align*}
    \mathbb{E}\left[ \hat{\*v}_{t+1} - \hat{\*v}_{t} | \mathcal{F}_t\right] = & \mathbb{E}\left[ \frac{1-\beta_2}{1-\beta_2^{t+1}} \left(\*g_t^2 - \frac{\*v_t}{1-\beta_2^t}  \right) \Big| \mathcal{F}_t\right] \\
= & \frac{1-\beta_2}{1-\beta_2^{t+1}}\mathbb{E}\left[ \*g_t^2 - \frac{\*v_t}{1-\beta_2^t} \Big| \mathcal{F}_t\right].
\end{align*}
Note that
\begin{align*}
    \mathbb{E}[\*v_t]
= \mathbb{E}\left[ (1-\beta_2)\sum_{j=1}^{t}\beta_2^{t-j}\*g_j^2 \right] = (1-\beta_2^t)\mathbb{E}[\*g_t^2].
\end{align*}
Push it back, we know for each $i\in[d]$,
\begin{align}
\label{equ:proof:zero_mean}
    \mathbb{E}\left[ \*e_i^\top \left( \hat{\*v}_{t+1} - \hat{\*v}_{t} \right) | \mathcal{F}_t\right] = 0.
\end{align}
On the other hand, for each $i\in[d]$,
\begin{align*}
    \left| \*e_i^\top \left( \hat{\*v}_{t+1} - \hat{\*v}_{t} \right)\right| = & \frac{1-\beta_2}{1-\beta_2^{t+1}}\left| \*e_i^\top \left( \*g_t^2 - \frac{\*v_t}{1-\beta_2^t} \right)\right|
\end{align*}
Note that both $\*e_i^\top\*g_t^2$ and $\frac{\*e_i^\top\*v_t}{1-\beta_2^t}$ is non-negative. Considering that
\begin{align*}
    \frac{\*e_i^\top\*v_t}{1-\beta_2^t} = \frac{1-\beta_2}{1-\beta_2^t}\sum_{j=0}^{t}\beta_2^{t-j}\*e_i^\top\*g_j^2 \leq G.
\end{align*}
And so
\begin{align}
\label{equ:proof:bounded_diff}
    \left| \*e_i^\top \left( \hat{\*v}_{t+1} - \hat{\*v}_{t} \right)\right| \leq \frac{1-\beta_2}{1-\beta_2^{t+1}}G \leq \frac{1-\beta_2}{1-\beta_2^{t_0}}G \leq \sqrt{2}(1-\beta_2)G,
\end{align}
where we apply the fact that $t>t_0$ and $t_0>\frac{\log\left( 1/2 \right)}{\log(\beta_2)}$.
Considering Equation~(\ref{equ:proof:zero_mean}) and (\ref{equ:proof:bounded_diff}), we know it is a martingale difference sequence. 
Now we apply the Azuma-Hoeffding Inequality \citep{wainwright2019high}, and get for any $i\in[d]$,
\begin{align*}
    \mathbb{P}\left[ \left| \sum_{k=t_0}^{t-1}\*e_i^\top \left( \hat{\*v}_{k+1} - \hat{\*v}_{k} \right) \geq c \right| \right] \leq & 2\exp\left(-\frac{c^2}{2\sum_{k=t_0}^{t-1}\left(\sqrt{2}(1-\beta_2)G\right)^2 }\right) \\
= & 2\exp\left(-\frac{c^2}{4G^2(1-\beta_2)^2(t-t_0)}\right).
\end{align*}
Set the R.H.S. as $\delta$, we obtain
\begin{align*}
    c = \sqrt{4G^2(1-\beta_2)^2(t-t_0)\log\left(\frac{2}{\delta}\right)}.
\end{align*}
Finally we get
\begin{align*}
    \|\hat{\*v}_t - \hat{\*v}_{t_0}\|_\infty < \sqrt{4G^2(1-\beta_2)^2(t-t_0)\log\left(\frac{2}{\delta}\right)},
\end{align*}
as desired. That completes the proof

\end{proof}

\end{document}